\documentclass[runningheads,a4paper]{llncs}

\usepackage{amssymb}
\usepackage{graphicx}
\usepackage{verbatim}
\usepackage{url}
\usepackage{xcolor}
\usepackage{floatrow}
\usepackage{multirow}
\usepackage{multimedia}
 
\begin{document}

\title{Concurrent Segmentation and Localization for Tracking of Surgical Instruments}
\titlerunning{CSL for Tracking of Surgical Instruments}

\newcommand\blfootnote[1]{%
  \begingroup
  \renewcommand\thefootnote{}\footnote{#1}%
  \addtocounter{footnote}{-1}%
  \endgroup
}

\author{Iro Laina\inst{1}*
\and  Nicola Rieke\inst{1}*
\and  Christian Rupprecht\inst{1,2}
\and  Josu\'{e} Page Vizca\'{i}no\inst{1}
\and Abouzar Eslami \inst{3}
\and  Federico Tombari\inst{1}
\and Nassir Navab\inst{1,2}}


\authorrunning{I. Laina and N.Rieke et al.} 

\institute{Computer Aided Medical Procedures (CAMP), TU Munich, Germany 
\and Johns Hopkins University, Baltimore, USA
\and Carl Zeiss MEDITEC M\"unchen, Germany \\
\email{Nicola.Rieke@tum.de}}


\maketitle             

\blfootnote{* I. Laina and N. Rieke contributed equally to this work.}

\begin{abstract}
Real-time instrument tracking is a crucial requirement for various computer-assisted interventions. 
In order to overcome problems such as specular reflections and motion blur, we propose a novel method that takes advantage of the interdependency between localization and segmentation of the surgical tool.
In particular, we reformulate the 2D instrument pose estimation as heatmap regression and thereby enable a concurrent, robust and near real-time regression of both tasks via deep learning. 
As demonstrated by our experimental results, this modeling leads to a significantly improved performance than directly regressing the tool position and allows our method to outperform the state of the art on a Retinal Microsurgery benchmark and the MICCAI EndoVis Challenge 2015.
\end{abstract}

\section{Introduction and Related Work}
\label{sec:introduction}
\setcounter{footnote}{0}
In recent years there has been significant progress towards computer-based surgical assistance in Minimally Invasive Surgery (MIS) and Retinal Microsurgery (RM). 
Two of the key components are segmentation and localization of surgical instruments during the intervention:
tool segmentation provides, for example, suitable regions for a graphical overlay of additional information without obstructing the surgeon's view;
tool movement is an indicator for surgical workflow analysis; localization of the instrument tips in RM allows proximity estimation to the retina by aligning a cross-sectional view.
For these tasks, marker-free approaches are particularly desirable as they do not interfere with the surgical workflow and they do not require modifications to the tracked instrument. 
Despite recent advances, the vision-based tracking of surgical tools in \emph{in-vivo} scenarios remains challenging, as summarized by Bouget \emph{et al.}~\cite{bouget2017vision}, mainly due to nuisances such as strong illumination changes and blur.  
Prior work in the field relies on handcrafted features, such as Haar wavelets~\cite{sznitman2013unified}, gradient~\cite{rieke2016realMEDIA,bouget2015detecting,li2014instrument,ye2016real} or color features~\cite{zhou2014visual,speidel2009automatic}, 
which come with their own advantages and disadvantages. 
While color features, for example, are computationally cheap, they are not robust towards strong illumination changes which are often present during the surgery.
Gradients features, on the other hand, are not reliable to withstand the typical motion blur of the tools. 
Rieke \emph{et al.}~\cite{rieke2016real} employed both feature types in two separate Random Forests and proposed to adaptively choose the more reliable one, depending on the confidence of the respective forest's leaf nodes. 
Since their explicit feature representation incorporates implicit simplifications, this tends to limit the generalization power of the forests and therefore leads to the risk of tracking failure during surgery. 
Furthermore, temporal trackers~\cite{rieke2016real,rieke2015,li2014instrument} require an initialization of the region of interest.
Sarikaya \emph{et al.}~\cite{SaCoGuTMItoappear} present a deep learning approach for tool detection via region proposals, which provides a bounding box and but not a precise localization of the landmarks. 
Instead of tracking the tool directly, two-step methods based on tool segmentation have also been proposed.
Color, HOG and SIFT features were employed by Allan \emph{et al.}~\cite{allan2013toward} for pixel-wise classification of the image. 
The position was subsequently determined based on largest connected components. 
Instead of reducing the region of interest, Reiter \emph{et al.}~\cite{reiter2012marker} employ the segmentation as a post-processing step for improving the localization accuracy. %
Recent segmentation methods~\cite{garcia2016real,pakhomov2017deep} can also be employed for these two-step approaches.
However, the observation that segmentation can be used during both pre- and post-processing suggests that tracking of an instrument landmark and its segmentation are not only dependent, but indeed interdependent.

\begin{figure}[t]
    \centering
        \includegraphics[width=1\textwidth]{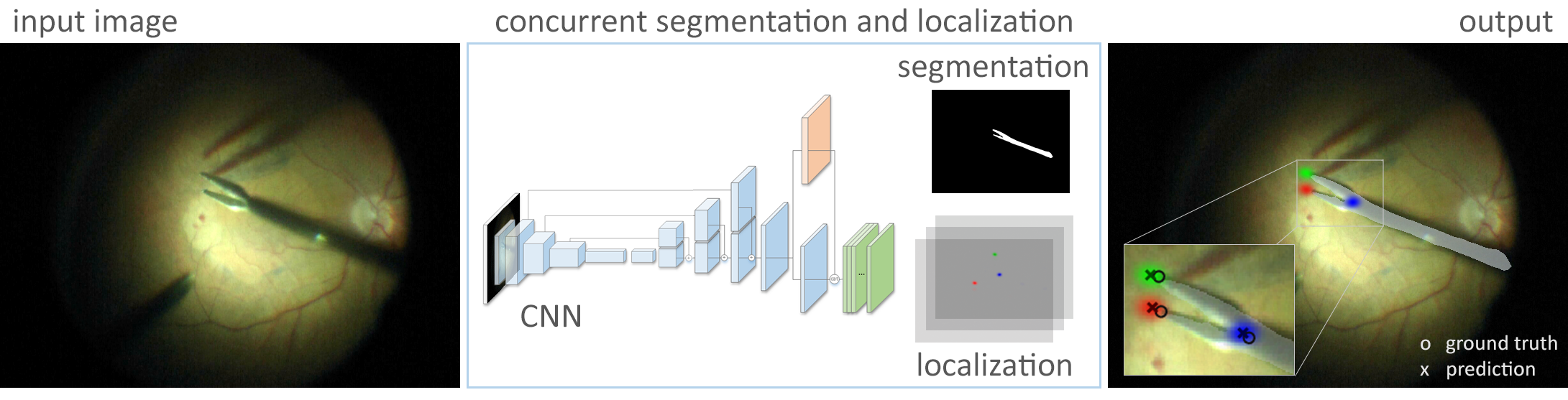}
      \caption{\textbf{Overview of the proposed method (CSL):} Concurrent semantic segmentation and landmark localization with a CNN-based approach. Formulating localization as regression of heatmaps allows training the model end-to-end with shared weights for both tasks and results in higher accuracy than regressing the 2D coordinates of the landmarks.}
\label{fig:teaser}
\end{figure}

Our contributions are as follows.
Instead of carrying out the tasks as two subsequent pipeline stages, we propose to perform tool segmentation and pose estimation simultaneously, in a unified deep learning approach (Fig.~\ref{fig:teaser}).
To this end, we reformulate the pose estimation task and model the problem as a heatmap regression where every pixel represents a confidence proportional to its proximity to the correct landmark location.
This modeling allows for representing semantic segmentation and localization with equal dimensionality, leveraging on their spatial dependency and facilitating simultaneous learning. 
It also enables employing state-of-the-art deep learning techniques, such as Fully Convolutional Residual Networks~\cite{laina2016deeper,he2016deep}.
The resulting model is trained jointly and end-to-end for both tasks.
It relies only on contextual information and is thus capable of reaching both objectives efficiently without requiring any post-processing technique.
We compare the proposed method to state-of-the-art algorithms on the EndoVis Challenge\footnote{MICCAI 2015 Endoscopic Vision Challenge Instrument Segmentation and Tracking Sub-challenge \url{http://endovissub-instrument.grand-challenge.org}} and on a benchmark dataset of \emph{in-vivo} RM sequences, on which we also outperform other popular CNN architectures, such as U-net~\cite{ronneberger2015u} and the FCN-based approach of \cite{garcia2016real}.
To the best of our knowledge, this is the first approach that employs deep learning for surgical instrument tracking and 2D pose estimation by predicting semantic segmentation and localization simultaneously and is successful despite limited data.

\section{Method}
\label{sec:method}
\begin{figure}[t]
	\centering
	\includegraphics[width=\textwidth]{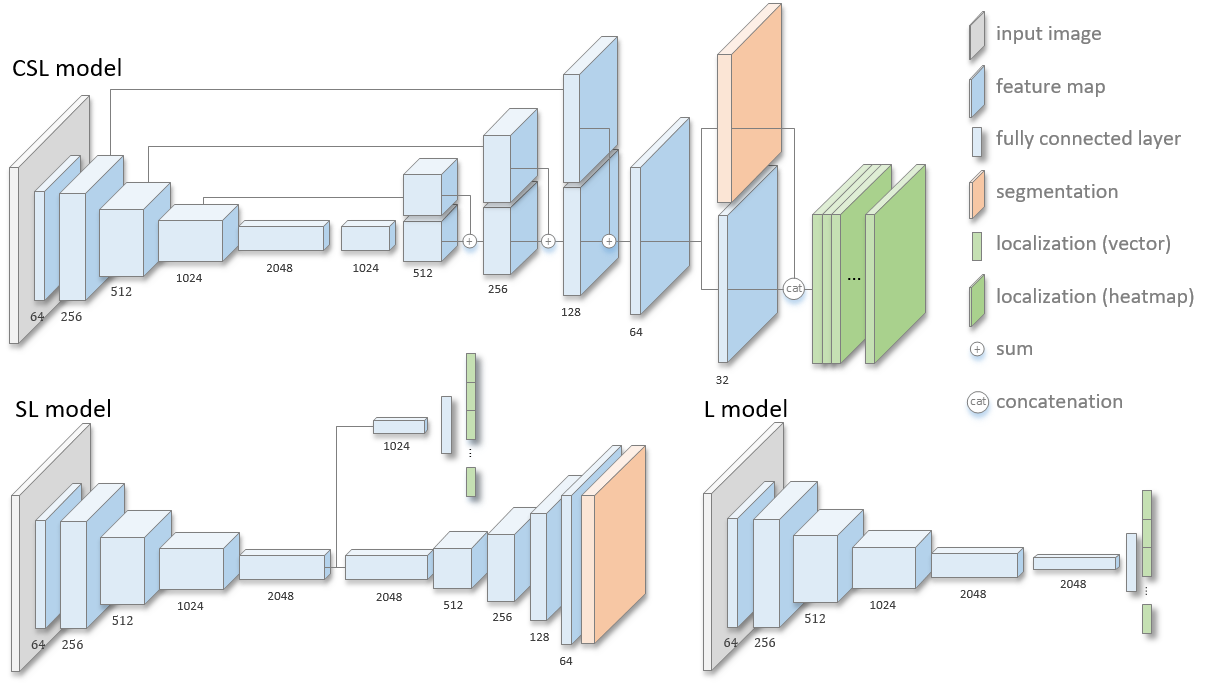}
	\caption{\textbf{Modeling Strategies:} The proposed CSL architecture and the two baseline models SL and L. }
	\label{fig:models}
\end{figure}

This section describes our CNN-based approach to model the mapping from an input image to the location of the tool landmarks and the corresponding dense semantic labeling. 
For this purpose, we motivate the use of a fully convolutional network, that models the problem of landmark localization as a regression of a set of heatmaps (one per landmark) in combination with semantic segmentation. 
This approach exploits global context to identify the position of the tool and has clear advantages comparing to patch-based techniques~\cite{alsheakhali2016detection}, which rely only on local information, thus being less robust towards false positives, e.g. specular reflections on the instrument or shadows.
We compare the proposed architecture and discuss its advantage over two baselines.
A common block for all discussed architectures is the encoder (Sec.~\ref{ssec::encoder}), which progressively down-samples the input image through a series of convolutions and pooling operations. The differences lie in the subsequent decoding stages (Sec.~\ref{ssec:decoder}) and the output formulation. 
An overview of these models is depicted in Fig.~\ref{fig:models}. 
We denote a training sample as $(X,S,y)$, where $y \in \mathbb{R}^{(n \times 2)}$ refers to the 2D coordinates of $n$ tracked landmarks in the image $X \in \mathbb{R}^{w\times h\times3}$, $S\in\mathbb{R}^{\frac{w}{2}\times\frac{h}{2}\times c}$ represents the semantic segmentation for $c$ labels and $w$, $h$ denote the image width and height respectively. 

\subsection{Encoder}
\label{ssec::encoder}
For the encoding part of the three proposed models, we employ ResNet-50~\cite{he2016deep}, a state-of-the-art architecture that has achieved top performance in several computer vision tasks, such as classification and object detection.
It is composed of successive \emph{residual} blocks, each one consisting of several convolutions and a shortcut (identity) connection summed to its output.
In this way, it allows for a very deep architecture without hindering the learning process and with relatively low complexity.
Although deeper versions of ResNet exist, we use the $50$-layer variant, as computation time is still crucial for our problem.
As input to the network, we consider images with $w = h = 480$ pixels. 
Thus, the feature maps at the last convolutional layer of ResNet have a resolution of $15\times15$ pixels. 
The last pooling layer and the loss layer are removed.

\subsection{Decoder Tasks}
\label{ssec:decoder}
We then define three different CNN variants, appended to the encoder, to find the best formulation for our task. In the following we outline the characteristics of each model, discuss their differences and motivate the choice of the final proposed model. 

\subsubsection{Localization (L):}
\label{ssec::L}
First, we examine the na\"ive approach, frequently used in literature~\cite{rupprecht2016sensor}, that regresses the real 2D locations of the landmarks directly as a $2 \times n$ dimensional vector representing the $x$ and $y$ coordinates of the $n$ tracked landmarks of the instrument.
Here, the segmentation task is excluded. %
To further reduce the spatial dimensions of the last feature maps, we append another residual block with stride to the end of the encoder ($8\times8\times2048$). Similarly to the original architecture~\cite{he2016deep}, this is followed by a $8\times8$ average pooling layer and a fully-connected layer with $2n$ units which produces the output. 
This dimensionality reduction is needed so that the averaging is not applied over a large region, which would result in a greater loss of spatial information, thus affecting the precision with which the network is able to localize.
In this case, the training sample is $(X,y)$ and the predicted location is $\tilde{y} \in \mathbb{R}^{2\times n}$. The network is trained with a standard $L_2$ loss:
$l_L(\tilde{y}, y) = || \tilde{y} - y||_2^2$.

\subsubsection{Segmentation and Localization (SL):}
\label{ssec::SL}
In this model, we regress the 2D locations and additionally predict the semantic segmentation map of an input within a single architecture. 
Both tasks share weights along the encoding part of the network and then split into two distinct parts to model their different dimensionality.
For the regression of the landmark positions we follow the aforementioned model (\emph{L}). For the semantic segmentation, we employ successive residual up-sampling layers as in \cite{laina2016deeper}, to predict the probability of each pixel belonging to a specified class, e.g. manipulator, shaft or background. 
Due to real-time constraints, we produce the network output with half of the input resolution and bilinearly up-sample the result.   
By sharing the encoder weights, the two tasks can influence each other while upholding their own objectives.
Here, the training sample is $(X,S,y)$, and the prediction consists of $\tilde{y} \in \mathbb{R}^{2\times n}$ and $\tilde{S} \in \mathbb{R}^{\frac{w}{2}\times \frac{h}{2} \times c}$.
The network is trained by combining the losses for the separate tasks:
$l_{SL}(\tilde{y}, y, \tilde{S}, S) =  \lambda_Ll_L(\tilde{y}, y) + l_S(\tilde{S}, S)$, where $\lambda_L$ balances the influence of both loss terms. For the segmentation we employ a pixel-wise softmax-log loss:
\begin{equation}
  l_S(\tilde{S}, S) = -\frac{1}{wh}\sum_{x=1}^w\sum_{y=1}^h\sum_{j=1}^c S(x,y,j) \log\left( \frac{e^{\tilde{S}(x,y,j)}}{\sum_{k=1}^ce^{\tilde{S}(x,y,k)}}\right)
\end{equation}

\subsubsection{Concurrent Segmentation and Localization (CSL):}
\label{ssec::CSL}
In both \emph{L} and \emph{SL} architectures, only a single 2D position is considered as the correct target for each landmark. 
However, manual annotations can differ in a range of several pixels, which in turn implies discrepancies or imprecise labeling. 
Predicting an absolute target location is somewhat arbitrary and ignores image context.
Therefore, in the proposed model (CSL), we address this problem by regressing a \textit{heatmap} for each tracked landmark instead of its exact coordinates, as recently used in the field of human pose estimation~\cite{pavlakos2017volumetric,pfister2015flowing}. 
The heatmap represents the confidence of being close to the actual location of the tracked point and is created by applying a Gaussian kernel to its ground truth position.
The heatmaps have the same size as the segmentation and can explicitly share weights over the entire network.
We further enhance the architecture with long-range skip connections that \textit{sum} lower-level feature maps from the encoding into the decoding stage, in addition to the residual connections of the up-sampling layers~\cite{laina2016deeper}. 
This allows higher resolution information from the initial layers to flow to the output layers without being compressed through the encoder, thus increasing the model's accuracy. 
Finally, we enforce a strong dependency of the two tasks by only separating them at the very end and concatenating the predicted segmentation scores (before softmax) to the last set of feature maps as an auxiliary means for guiding the location heatmaps.
The overall loss is given by:
\begin{equation}
  l_{CSL} = l_S(\tilde{S}, S) + \frac{\lambda_H}{n}\sum_{i=1}^n\sum_{x=1}^w\sum_{y=1}^h ||\frac{1}{\sqrt{2\pi\sigma^2}}e^{-\frac{|| y_i - (x,y)^T  ||_2^2}{2\sigma^2}} - \tilde{y}^*_{x,y,i}||_2^2
\end{equation}
The standard deviation $\sigma$ controls the spread of the Gaussian around the landmark location $y_i$. In testing, the point of maximum confidence in each predicted heatmap $\tilde{y}^*_i \in \mathbb{R}^{\frac{w}{2}\times\frac{h}{2}\times n}$ is used as the location of the instrument landmark.
Notably, a misdetection is indicated by high variance in the predicted map.

\section{Experiments and Results}

\label{sec::experiments}
In this section, we evaluate the performance of the proposed method in terms of localization of the instrument landmarks, as well as segmentation accuracy.
\\

\textbf{Datasets:}
The \texttt{Retinal Microsurgery} dataset~\cite{rieke2016realMEDIA} consists of $18$ \emph{in-vivo} sequences, each with $200$ frames of resolution $1920~\times~1080$ pixels.
The dataset is further classified into four instrument-dependent subsets.
The annotated tool joints are $n=3$ and semantic classes $c=2$ (tool and background).

In the \texttt{EndoVis} challenge, the training data contains four \emph{ex-vivo} 45s sequences and the testing includes the rest 15s of the same sequences, plus two new 60s videos.
Notably, the guidelines require to exclude the respective surgery for training when testing on the additional 15s sequence and one of the long testing sequences include a previously unseen tool type.
All sequences have a resolution of $720 \times 576$ pixels and include one or two surgical instruments. 
There is $n=1$ joint per tool and $c=3$ semantic classes (manipulator, shaft and background).
\\

\textbf{Implementation details:}
\label{ssec::implementation}
The encoder is initialized with ResNet-50 weights pretrained on ImageNet.
All newly added layers are randomly initialized from a normal distribution with zero mean and $0.01$ variance.
All images are resized to $640\times480$ pixels and augmented during training with random rotations $[-5^{\circ},5^{\circ}]$, scaling $[1,1.2]$, random crops of $480 \times 480$, gamma correction with $\gamma \in [0.9,1.1]$, a multiplicative color factor $c \in [0.8,1.2]^{3}$ and specular reflections. 
For localization, we set $\sigma = 5$ for \texttt{RM} and $\sigma = 7$ for \texttt{EndoVis} in which the tools are larger.
All CNNs are trained with stochastic gradient descent with learning rate $10^{-7}$, momentum $0.9$ and empirically chosen $\lambda_{L},\lambda_{H}=1$. 
The inference time is 56ms per frame on a NVIDIA GeForce GTX TITAN X using MatConvNet.

\subsection{Evaluation of Modeling Strategies}
\label{ssec:architecture}

\begin{figure}[t]
	\centering
	\includegraphics[width=0.95\textwidth]{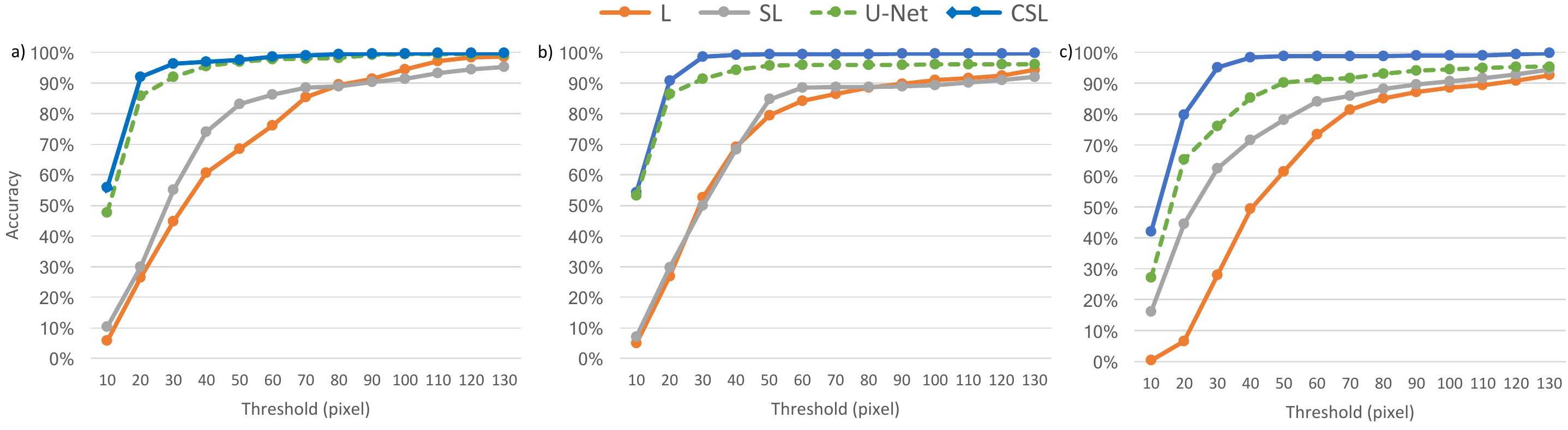}
	\caption{\textbf{Evaluation of Modeling Strategies:} Accuracy of the models by means of Threshold Score for the left tip (a), right tip (b) and center joint (c) of the instrument. }
	\label{fig:exp_models}
\end{figure}

\begin{figure}[!htp]
	\centering
	\includegraphics[width=\textwidth]{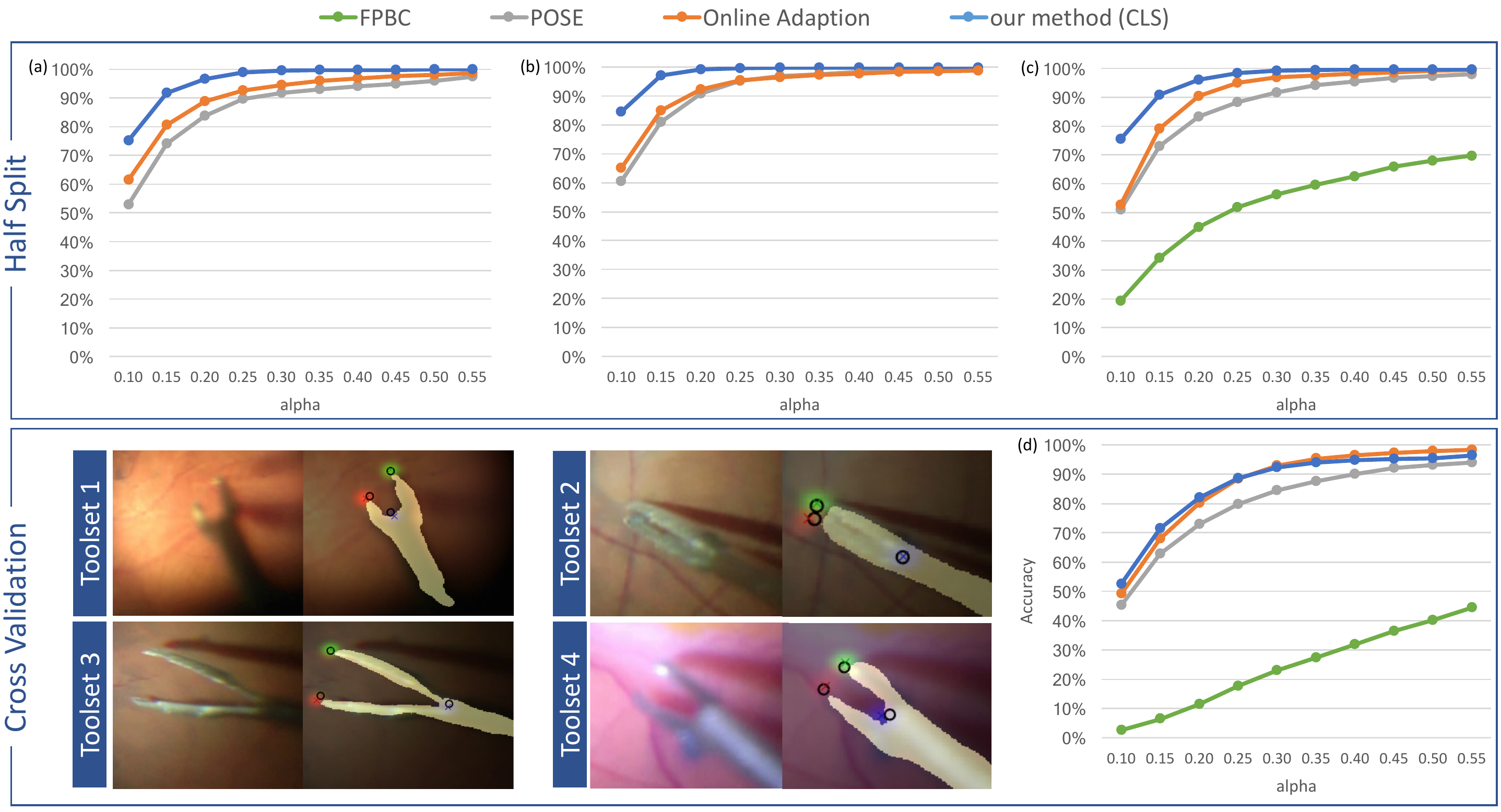}
	\caption{\textbf{RM dataset:} Comparison to FPBC~\cite{sznitman2014fast}, POSE~\cite{rieke2016realMEDIA} and Online Adaption~\cite{rieke2016real}, measured by the metric \emph{KBB}. The charts (a) to (c) show the accuracy for the left tip, right tip and center joint, respectively, for the \emph{Half Split} experiment. In the \emph{Cross Validation}, the training set is given by $3$ instrument dependent subsets and the method is tested on the remaining set. (d) shows the average \emph{KBB} score for the center point.}
	\label{fig:exp_RM}
\end{figure}

First, we evaluate the models for tool landmark localization by training on $9$ sequences of the RM dataset and testing on the remaining ones.
In Fig.~\ref{fig:exp_models}, the baseline of explicitly predicting the 2D coordinates of the landmark locations (\textbf{L}) shows the lowest accuracy, while after combining localization with the segmentation task (\textbf{SL}) we observe increased performance. 
The proposed \textbf{CSL} model achieves the highest accuracy of over 90\% for both tool tips and 79\% for the center joint considering an acceptance pixel threshold of 20 pixels. 
Our model exploits contextual information for precise localization of the tool, by sharing features with the semantic segmentation task. 
Another baseline is the U-Net architecture~\cite{ronneberger2015u} trained with the same objectives. CSL is consistently more accurate for the localization task, as well as for the segmentation, achieving a DICE score of \textbf{75.4\%}, comparing to 74.4\% for CSL without the skip connections, 73.7\% for SL and 72.5\% for U-net.

\subsection{Retinal Microsurgery}
\label{ssec::RM}
Analogously to~\cite{rieke2016realMEDIA}, we train the proposed model (CSL) on all first halves of the $18$ RM sequences and evaluate on the remaining frames, referred to as \emph{Half Split Experiment}.
As shown in Fig.~\ref{fig:exp_RM}, the proposed method clearly outperforms the state-the-art-methods, reaching an average accuracy of more than 84\% for the \emph{KBB} score~\cite{rieke2016realMEDIA} with $\alpha = 0.15$.
In a second experiment, we evaluate the generalization ability of our method not only to unseen sequences and but also to unknown geometry. We employ a leave-one-out scheme on the subsets given by the $4$ different instrument types, referred to as \emph{Cross Validation Experiment}, and show that our method achieves state-of-the-art performance.

\subsection{EndoVis Challenge}
\label{ssec::Endovis}
\begin{figure}[]
  \centering
  \begin{minipage}[b]{0.45\textwidth}
    \includegraphics[width=\textwidth]{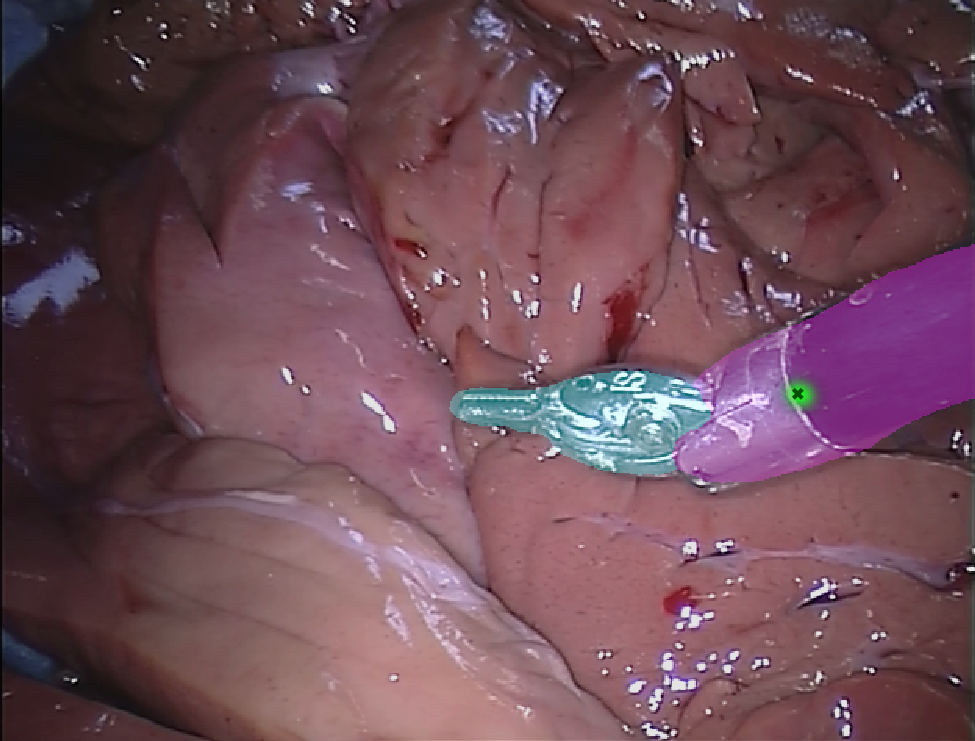}
  \end{minipage}
  \hfill
  \begin{minipage}[b]{0.45\textwidth}
    \includegraphics[width=\textwidth]{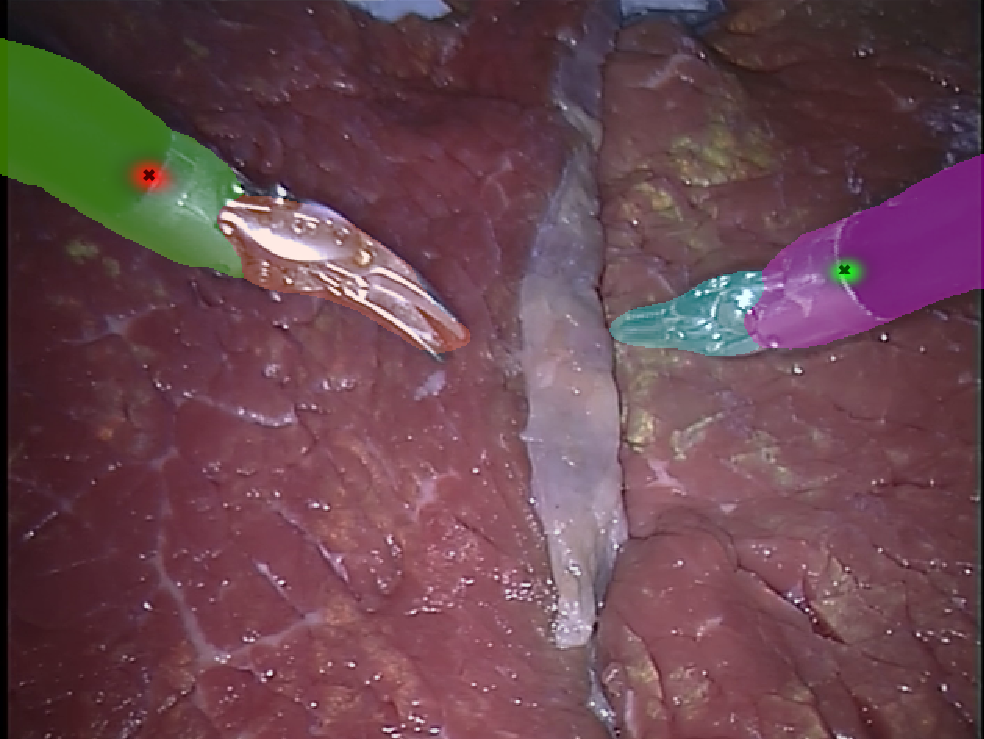}
  \end{minipage}
  \caption{\textbf{Qualitative Result (EndoVis):} The number of semantic segmentation classes and tracked joints can easily be adjusted via the number of heatmaps in our method. Left: Only one instrument is present. Right: Two instruments are present and can even be distinguished as two separate tools. Notably, the left instrument was not seen in the training dataset.  }
  \label{fig:Endovis}
\end{figure}

For this publicly available dataset, we performed our experiments in a leave-one-surgery-out fashion, as specified by the guidelines.
We report our quantitative results in Table~1 and compare to the previous state of the art, which we significantly outperform.  
In all of our experiments, the network was trained with the objective of multi-class segmentation. For the binary prediction, the instrument classes (Shaft and Grasper) were merged.
Notably, the proposed method can also distinguish among parts of multiple instruments (Fig.~\ref{fig:Endovis}), for example left and right, when trained with $c=5$ classes (left shaft, left grasper, right shaft, right grasper, background) and $n=2$ joints.

A challenging aspect of this dataset is that two instruments can be present in the testing set, while only one is included in the training. 
To alleviate this problem, we additionally augment with horizontal flips, such that the instrument is at least seen from both sides. 
Moreover, in Sets 5 and 6 the network was capable of successfully localizing and segmenting a previously unseen instrument and viewpoint\footnote{The challenge administrators believe that the ground truth regarding tracking for sequence $5$ and $6$ is in fact not as accurate as for the rest of the sequences, which explains the higher localization errors.}.

\begin{table}
  \scalebox{0.75}{
  \begin{tabular}{  l r  r  r  r | r  r  r  r |  r r r r |r}
      \hline
     & \multicolumn{4}{c}{Binary}  &  \multicolumn{4}{c}{Shaft} &   \multicolumn{4}{c}{Grasper} & Joint  \\

 Sequence   & \multicolumn{2}{r}{B.Acc.}     Rec. &  Spec. &  DICE & Prec. & Rec. &  Spec. & DICE & Prec. & Rec. &  Spec. & DICE & loc.~error  \\ \hline
   1 & 91.9	 & 85.0	 & 98.7 & 	 88.5 & 91.6 & 79.2 & 99.1  & 88.9 & 71.6 & 76.2 & 98.7 & 73.8 & 39.0/30.8\\
2& 94.8	 & 90.0	 & 99.7 & 	 93.0 & 96.5  & 90.9 & 99.8  & 93.6& 90.7 & 82.0 & 99.8 & 86.1 & 9.7\\
3 & 94.7	 & 90.1	 & 99.3 &  91.6 & 94.7 & 89.1 & 99.5  & 91.5 & 87.0 & 86.8 & 99.7 & 86.7 & 10.9\\
4& 91.1	 & 83.1	 & 99.0 &  85.8 & 89.0 & 82.9 & 99.2  & 85.3 & 74.9 & 65.4 & 99.6 & 68.4 & 13.0\\
5 & 91.5	 & 84.2	 & 98.8 &  87.3 & 90.1 & 82.8 & 99.1  &  86.2 & 79.2 & 75.9 & 99.2 & 77.1 & 38.4/60.0\\
6 & 91.7	 & 84.9	 & 99.0 &  88.9 & 92.5 & 78.0 & 99.3  & 84.5 & 71.1 & 78.1 & 98.4 & 74.1 & 36.4/63.9\\
   \hline 
   CSL (mean) & \textbf{92.6} & 86.2	 & \textbf{99.0} &  88.9 & 92.4 & 83.8 & 99.3 & 87.7 & 79.1 &77.4 & 99.2 & 77.7 &24.8/51.6\\ \hline \hline
   FCN~\cite{garcia2016real} & 83.7 & 72.2 & 95.2 & - & - & - & - &- &- & - &- &- &-\\
   FCN+OF~\cite{garcia2016real} & 88.3 & \textbf{87.8} & 88.7& - & - & - &- &- &- &- & - & - &-\\
   DRL~\cite{pakhomov2017deep} & 92.3 & 85.7 & 89.8 & - & - & - &- &- &- &- & - & - &-\\
   \hline
   \multicolumn{14}{l}{Balanced Accuracy (\textit{B.Acc.}), Recall  (\textit{Rec.}), Specificity (\textit{Spec.}), DICE and Precision (\textit{Prec.}) are in  \%}\\
   \multicolumn{14}{l}{The average localization error (\textit{loc.~error}) is in pixel.}\\
   \caption{\textit{Cross-Validation} results for \texttt{EndoVis}.}
\end{tabular}

 }
\end{table}

\section{Conclusion}
\label{sec:conclusion}
In this paper, we propose to model the localization of surgical instrument landmarks as heatmap regression.
This allows us to leverage deep-learned features via a CNN to concurrently predict the instrument segmentation and its articulated 2D pose in an end-to-end manner.
It is worth noting that the resulting method is flexible regarding the number of tracked joints and semantic classes and even allows to distinguish between left and right instrument. These objectives can be specified during training by simply setting the number of the respective semantic classes and heatmaps. 
The inference time is near real-time and the method does not require an initialization, post-processing technique or temporal regularization.
The performance is evaluated on two different surgical intervention benchmarks, on which the proposed approach delivers state-of-the-art results.

\bibliography{references}
\end{document}